\documentclass[preprint,11pt,nonatbib]{elsarticle}
\journal{arXiv.org}

\usepackage{newtxtext}
\usepackage{newtxmath}
\usepackage{bm}

\usepackage[a4paper]{geometry}
\usepackage{setspace}
\makeatletter
\let\c@author\relax
\makeatother
\usepackage[backend=biber, style=authoryear, 
maxcitenames=2, maxbibnames=99, giveninits=true, hyperref=true,
natbib=true, uniquelist=false, uniquename=false]{biblatex}
\DeclareNameAlias{sortname}{family-given}
\usepackage{tablefootnote}
\usepackage{array}
\usepackage{booktabs}
\usepackage{amsmath}
\usepackage{amstext}
\usepackage{graphicx}
\usepackage{lineno}
\usepackage{siunitx} 

\usepackage{hyperref}
\usepackage{xurl}
\hypersetup{
  colorlinks=true,
  linkcolor=blue,
  citecolor=blue,
  urlcolor=blue
}

\usepackage{enumitem}

\begin{document}

\begin{frontmatter}

\title{Hierarchical forecasting with a top-down alignment of independent level forecasts}

\author[mainaddress]{Matthias Anderer} \ead{matthias@matthiasanderer.de}
\author[secondaryaddress]{Feng Li\corref{correspondingauthor}} \ead{feng.li@cufe.edu.cn}
\address[mainaddress]{Matthias Anderer GmbH, Holzkirchen, Germany}
\address[secondaryaddress]{School of Statistics and Mathematics, Central University of
  Finance and Economics, Beijing 102206, China.}
\cortext[correspondingauthor]{Corresponding author.}

\begin{abstract}
  Hierarchical forecasting with intermittent time series is a challenge in both research and empirical studies. Extensive research focuses on improving the accuracy of each hierarchy, especially the intermittent time series at bottom levels. Then hierarchical reconciliation could be used to improve the overall performance further.  In this paper, we present a \emph{hierarchical-forecasting-with-alignment} approach that treats the bottom level forecasts as mutable to ensure higher forecasting accuracy on the upper levels of the hierarchy. We employ a pure deep learning forecasting approach N-BEATS for continuous time series at the top levels and a widely used tree-based algorithm LightGBM for the intermittent time series at the bottom level. The \emph{hierarchical-forecasting-with-alignment} approach is a simple yet effective variant of the bottom-up method, accounting for biases that are difficult to observe at the bottom level.  It allows suboptimal forecasts at the lower level to retain a higher overall performance.  The approach in this empirical study was developed by the first author during the M5 Forecasting Accuracy competition, ranking second place.  The method is also business orientated and could benefit for business strategic planning.
\end{abstract}


\end{frontmatter}


\newpage
\setcounter{page}{1}

\section{Introduction}
\label{sec:intro}

With the rise of business oriented time-series data collection across industries,
improving forecasting performance becomes vital for business success. Among the vast
forecasting techniques, conventional statistical forecasting methods are dominant in the
forecasting domain (for a survey in this area see \citet{hyndman2018forecasting}, while
for an encyclopedia review of forecasting theory and applications, see
\citet{petropoulos2021forecasting}). Machine learning algorithms are widely used in time
series forecasting, notably long short term memory (LSTM) neural networks and their
variations, deep neural architectures such as N-BEATS \citep{oreshkin2019n}, or the
Temporal Fusion Transformer \citep{lim2019temporal}. The top-ranking method in the M4
competition \citep{smyl2020hybrid} suggested that hybrid models combining statistical
approaches and machine learning methods could further improve the forecasting accuracy.

Although statistical approaches tend to be more interpretable, a cost-efficient forecast
usually requires expert information in e.g., features engineering and model
specification. Those approaches work well with individual time series, but it is hard to
scale them up with sizeable related time series sets. On the contrary, industrial machine
learning platforms with pipelines provided for global forecasting solutions do not
necessarily achieve the best optimality without extensive tuning. Still, they are easy to
scale up for massive amounts of time-series data. It is thus appealing for industrial
forecasters to adopt machine learning approaches.

Intermittent time series, which contains many zeros in the data, is a typical case in
demand forecasting. Such time series are difficult to forecast because it is toilsome to
make realistic assumptions across all the time series horizons. Dating back to the 1970s,
researchers observed that conventional time series methods, such as exponential smoothing,
did not perform well for forecasting intermittent demand time series
\citep{croston1972forecasting}. A straightforward approach, known as the Croston's method,
was proposed by splitting the time series into the non-zero component and stochastic
component. The two parts were then forecasted separately and the final forecast was
factorized by the two individual forecasts in the end. See e.g.,
\citet{croston1972forecasting,rao1973comment,shenstone2005stochastic} for the related
study. Although such a method has been proven to be helpful, it often produces forecast
bias \citep{leven2004inventory}. Many variants were proposed to improve the accuracy
\citep{shenstone2005stochastic,kourentzes2021elucidate}, or reduce the bias
\citep{syntetos2005accuracy,nikolopoulos2011aggregate} of Croston's method.

Attracted by the nonparametric and assumption-free nature of neural networks, researchers
shift the focus to neural networks \citep{caner2020intermittent}, and directly construct
neural networks for time series with particular patterns, most notably as short
intermittent time series \citep{kourentzes2013intermittent} and long lumpy time series
\citep{gutierrez2008lumpy}.  One widely known approach is DeepAR - an architecture based
on autoregressive recurrent networks \citep{salinas2020deepar}.

Another challenge in the supply chain or online sales industry is simultaneously
forecasting an extensive collection of time series. Such time series collection usually
involves a hierarchical structure from different cross-sectional levels. For example, the
dataset in the M5 competition \citep{makridakis2021m5} consists of 3049 products, which
are sold across ten stores located in three states in the U.S. The products are further
classified into three product categories and seven product departments. Notably,
information from different cross-sectional levels may be crucial for improving overall
forecasting performance. Although the practitioners find incorporating the cross-sectional
information improves the overall forecasting performance, a challenge exists that the
forecasts on each hierarchy should be coherent with the aggregation structure of the
collection of time series. A hierarchical forecasting method could utilize the structural
information to improve the overall forecasting performance and retain the forecast
coherence. See \citet{hyndman2018forecasting} for an introduction to hierarchical
forecasting.

A common phenomenon of hierarchical time series sets in the sales industry is that the
lowest level of the hierarchy exhibits a strong intermittent pattern, and the upper
hierarchy levels contain forecastable components such as the trend or seasonality
aggregated by the bottom level. Early research shows that spatial aggregation for
intermittent time series improves the overall forecast accuracy
\citep{zufferey2016forecasting}. Similar results are also found with temporal aggregated
intermittent time series in e.g., \citet{nikolopoulos2011aggregate} and
\citet{kourentzes2021elucidate}.

To this end, we consider a typical business planning scenario. Under certain
circumstances, managers at the headquarter or the investors may not be interested in the
forecasts of a particular product in a specific store, but they are more concerned with
the state level's forecasts for all products, which affect the overall revenue. By further
incorporating the product information as explanatory variables into the bottom level
models, we argue that decision-makers or business planners may benefit by further
forecasting performance improvement, especially at the top levels, from hierarchical
forecasting with intermittent time series, than by merely using individual forecasting
models for the bottom level intermittent data.

In this paper, we describe and analyze an approach proposed by the first author, who
utilizes standard machine-learning toolchains with a careful forecasting alignment scheme
achieving the second rank of the M5 Forecasting Accuracy competition. Unlike the
forecasting reconciliation approach that designs optimal proportions for the aggregations
to guarantee the coherence across multi-hierarchy forecasts, the proposed
\emph{hierarchical-forecasting-with-alignment} approach adjusts the sum of the forecasts
of the bottom level based on those aggregated at the top levels by selecting a tuning
parameter at the bottom level model so that the sum of the forecasts produced across
multiple forecasting horizons are aligned. Our analysis finds it particularly useful when
there is difficulty in finding an optimal reconciliation as suggested in e.g.,
\citet{wickramasuriya2019optimal}. A vital feature of this approach is that it improves
the overall forecast accuracy by allowing some suboptimal forecasts at the lower level but
retaining high forecasting accuracy at the top levels.

We organize the remainder of the paper as follows. Section \ref{sec:preliminary} presents
the preliminary for models used in our paper. The methodology is described in Section
\ref{sec:methodology}. Section \ref{sec:application} describes the implementation details
for the M5 competition. An ex-post analysis is provided in Section \ref{sec:postmortem}
and Section \ref{sec:conclusion} concludes the paper.

\section{Preliminary}
\label{sec:preliminary}

This section briefly describes two machine learning algorithms, i.e., N-BEATS
\citep{oreshkin2019n} and LightGBM \citep{ke2017lightgbm}, that are used for the proposed
\emph{hierarchical-forecasting-with-alignment} framework. Both methods are robust and
widely used in the machine learning community.

The N-BEATS \citep{oreshkin2019n} is a pure deep learning approach with a deep neural
architecture based on backward and forward residual links. It utilizes a very deep stack
of fully-connected layers that consist of $30$ stacks of depth $5$ with a total depth of
$150$ layers. The N-BEATS constructs a doubly residual stacking architecture with two
residual branches. One residual branch is running over the backcast prediction of each
layer, and the other one is running over the forecast branch of each layer. This
architecture improves the interoperability of deep learning models by especially allowing
for trend and seasonality decomposition. A desired feature for practitioners is that the
N-BEATS model does not require massive data sources, nor complex data transformation or
feature engineering, making the model extremely easy to use and deploy in the industry.

The performance of the N-BEATS model relies on the ensembling step, a forecasting
combination technique that combines the forecasts from different models. The final
forecasts provided by the N-BEATS are based on multi-scale aggregation with the median of
bagging selected models. The model pool for ensembling consists of N-BEATS models fitted
on one or several of the sMAPE, MASE and MAPE metrics and a selection of different window
lengths e.g., $2\times h$, $3\times h$, ... , $7\times h$, respectively, where $h$ is the
forecast horizon.

The N-BEATS model has already been proved to be robust and implemented on industry-level
platforms. We use the \texttt{GluonTS}' implementation \citep{gluonts2020} in this
analysis. The notable difference between \texttt{GluonTS}' implementation and the original
N-BEATS is splitting the training and forecasting series.  The N-BEATS picks a random
forecast point from the historical range of length $L_h$ for each selected time series,,
immediately preceding the last point in the train part of the time series. This method
requires further tuning of the hyperparameter $L_h$. However, \texttt{GluonTS} does not
use the $L_h$ parameter and cuts time windows by randomly selecting a time series and a
starting point on that time series, which reduces the computation burden.


The LightGBM \citep{ke2017lightgbm} is also a widely used machine learning algorithm based
on gradient boosting decision trees. The decision tree algorithm was developed for machine
learning tasks such as regression and classification. It partitions the input variables
into tree structures, and the final prediction is decided based on the tree of input
variables. The decision tree is a weak learner due to its simple decision
strategy. Gradient boosting algorithms are then used to ensemble the predictions in a
step-wise fashion to improve the prediction performance. The LightGBM further enhances the
efficiency and scalability of gradient boosting libraries, such as XGBoost
\citep{chen2016xgboost}, for large datasets with high-dimensional features.

\section{Methodology}
\label{sec:methodology}

\subsection{Improved N-BEATS ensembles for upper levels}

Hierarchical forecasting aims to build suitable forecasting models at different levels and
prevent overfitting at all levels using the hierarchical structure. We first train the
N-BEATS model using the top five levels data.  Although the data from the top five levels
are used to train the N-BEATS model, only the forecasts at the top level are employed for
the hierarchical alignment described in Section \ref{sec:alignment}. The other four levels
were served as a cross check.

During the M5 competition, the first author experimented with the N-BEATS ensembles for
the top-level time series during the validation time frame with different epochs.  It is
observable that the N-BEATS model started to overfit after approximately 12 epochs. For
that reason, N-BEATS ensembles for the top-level forecast with $10$ epochs were chosen for
the final evaluation models. A comparison with other single forecasting models on the top
level was not made due to the length of the competition time and the limited computing
resources available on the Kaggle platform. But according to the reported score by the
competitors \footnote{Available online at
  \url{https://www.kaggle.com/c/m5-forecasting-accuracy/discussion/134712}} on the Kaggle
platform, single models like the LightGBM at the top levels do not achieve better accuracy.

The stochastic gradient descent (SGD) algorithm used in the N-BEATS model has weak
learning stability. The learning accuracy is very sensitive to the learning rate, and the
convergence rate is slow.  The \emph{lookahead} optimizer \citep{zhang2019lookahead} is
adopted to the N-BEATS model trainer further to improve the training accuracy and learning
stability. Unlike the default stochastic searching scheme used in the SGD, the
\emph{lookahead} algorithm chooses a parameter search direction by looking ahead at the
sequence of weights generated by the standard optimizer provided by the \texttt{GluonTS}
package.

\subsection{Bias-adjusted LightGBM model for the bottom level}
\label{sec:lightgbm}

At the bottom level, we use a bias-adjusted LightGBM to model the intermittent time
series. The root mean squared error (RMSE) loss is used in the LightGBM as follows,
\begin{align*}
  RMSE = \sqrt{\frac{1}{h}\sum_{t = n + 1}^{n + h}(Y_t - \hat Y_t)^2}.
\end{align*}
where $Y_t$ is the true value, $\hat Y_t$ is the forecast, $h$ is the forecasting horizon,
and the residual is defined as $e_t = Y_t - \hat Y_t$. We further use a customized
gradient for the RMSE loss as follows
\begin{align}
 \label{eq:gradient}
  \mathrm{gradient} =  \begin{cases}
    -2 e_t     & e_t < 0, \\
    -2 \lambda   e_t & e_t \geq 0
  \end{cases}
\end{align}
where $\lambda > 0$ is a tuning parameter named the loss multiplier to allow for an
asymmetric loss. When $e_t \geq 0$, (i.e., the forecast is lower than the true value), the
loss multiplier magnifies ($\lambda > 1$) or minifies ($0 < \lambda < 1$) the gradient of
the loss function during the learning process. When $\lambda = 1$, the customized gradient
reduces to the conventional symmetric gradient. To this end, the loss multiplier controls
the bias during the training process to achieve a bias-variance trade-off effect for the
bottom level model. The customized gradient with the tuning parameter is particularly
useful for the final hierarchical alignment step described in Section
\ref{sec:alignment}. The corresponding Hessian is
\begin{align*}
  \mathrm{Hessian} =  \begin{cases}
    2    & e_t < 0, \\
    2  \lambda & e_t \geq 0.
  \end{cases}
\end{align*}

We independently train separate models for the time series data at each store for the
parallelization simplicity on the Kaggle computing platform. It is worth mentioning that
no rolling or lagged demand features were used in the LightGBM training, in contrast to
conventional statistical forecasting approaches. The motivation is to focus on the
intermittent characteristics at the bottom level. We argue that the intermittent demand
can be significantly influenced by factors like prices, time in a year, or special events
rather than the historical demand information. This approach with non-time series features
consistently achieves more stable forecast results. For the final forecasts, five
different multipliers around the optimal loss multiplier were used to build five
independent forecast models. The final forecasts at the bottom level are based on the mean
ensemble of the five models. See Section \ref{sec:application} for the implementation
details.

During the M5 competition, the first author also investigated the DeepAR
\citep{salinas2020deepar} approach for modeling the intermittent time series. Nonetheless,
the experiments found it difficult to achieve better results than LightGBM models or
obtain optimal reconciliation results compared with the minimum trace (MinT) optimal
reconciliation approach proposed in \citet{wickramasuriya2019optimal}.

\subsection{Aligning top level forecasts and aggregated bottom level forecasts}
\label{sec:alignment}

Given the independent forecasting results for the top level from stable N-BEATS
forecasting models with the number of epochs = 10, the final step is to find the optimal
bottom level forecasts produced by the LightGBM model. We tune the loss multiplier
($\lambda$) introduced for the LightGBM model in Section \ref{sec:lightgbm} so that the
RMSE between N-BEATS forecasts and aggregated bottom level forecasts reaches a
minimum. This step is called the hierarchical alignment, and the corresponding alignment
metric is defined in \autoref{eq:top-bottom-alignment} as
\begin{equation}
  \label{eq:top-bottom-alignment}
  \mathrm{arg\, min}_{\lambda} \left\{
    \sqrt{\frac{1}{h}\sum_{t = n + 1}^{n + h}\left( \hat Y_t^{(\mathrm{top})} - {Agg}(\hat Y_t^{(\mathrm{bottom})}(\lambda))\right)^2}
  \right\},
\end{equation}
where ${Agg}(\cdot)$ is the aggregating method used at the bottom level, and we use the
mean aggregation function throughout the M5 competition.

The hierarchical forecasting with alignment method is simple and applicable to any model
appropriate to the bottom level. The loss multiplier $\lambda$ in
\autoref{eq:top-bottom-alignment} is selected so that the sum of the forecasts produced
across the entire forecasting horizon are aligned. It means that the objective function
for the alignment depends on the forecasting horizon. During the M5 competition, a manual
grid search method was used to find the optimal loss multiplier for the LightGBM at the
bottom level, which can easily be automated in an industrial setting. This hierarchical
alignment method only requires running the model for the top level once. The optimal loss
multiplier could be obtained during the optimization stage of the bottom level model by
examining the alignment metric in \autoref{eq:top-bottom-alignment}. Note that in the
evaluation time frame during the M5 competition, \autoref{eq:top-bottom-alignment} is used
to determine optimal multiplier values because the actual future values are unknown. We
further provide an ex-post analysis in Section \ref{sec:postmortem} using the WRMSSE
metric (\autoref{eq:wrmsse}) when all data and results are disclosed after the
competition.

It is worth mentioning that there are alternative ways for hierarchical alignment. In
principle, this hierarchical alignment method should work for aligning all upper levels
with the bottom level. The reason that we only align with the top level is twofold. First,
the scoring metric in the M5 competition gives higher weights on upper levels because they
contain fewer series than the bottom level. Second, the computational cost on the top
level is much lower than aligning with several levels.

\section{Implementation details}
\label{sec:application}

The hierarchical forecasting with alignment approach has been successfully applied to the
M5 competition dataset, consisting of a hierarchical structure of daily sales data of
total $42,840$ series spanning 1,941 days. \autoref{tab:ts-hierarchy} depicts the
hierarchical structure of the data and the number of series per aggregation level.  One
could observe from \autoref{tab:ts-hierarchy} that the upper levels contain much less
series than the lower level.

\begin{table}
  \centering
  \caption{Overview of series per level and contribution to error metric for that level.
    Note that all hierarchical levels are equally weighted in the M5 competition.}
  \label{tab:ts-hierarchy}
  \resizebox{\textwidth}{!}{
    \begin{tabular}{clr}
      \toprule
      Hierarchy level & Description                                                          & Number of series \\
      \midrule
      1               & All products, all stores, all states                                 & 1                \\
      2               & All products by states                                               & 3                \\
      3               & All products by store                                                & 10               \\
      4               & All products by category                                             & 3                \\
      5               & All products by department                                           & 7                \\
      6               & Unit sales of all products, aggregated for each State and category   & 9                \\
      7               & Unit sales of all products, aggregated for each State and department & 21               \\
      8               & Unit sales of all products, aggregated for each store and category   & 30               \\
      9               & Unit sales of all products, aggregated for each store and department & 70               \\
      10              & Unit sales of product \emph{x}, aggregated for all stores/states     & 3,049            \\
      11              & Unit sales of product \emph{x}, aggregated for each State            & 9,147            \\
      12              & Unit sales of product \emph{x}, aggregated for each store            & 30,490           \\
\midrule
      Total           &                                                                      & 42,840            \\
      \bottomrule
    \end{tabular}
  }
\end{table}

The Root Mean Squared Scaled Error (RMSSE), which is a variant of the well-known Mean
Absolute Scaled Error (MASE) \citep{hyndman2006another}, is used for calculating the
out-of-sample forecasting error for each series as described in \autoref{eq:rmsse},
\begin{equation}
  \label{eq:rmsse}
  RMSSE = \sqrt{\frac{1}{h}\frac{\sum_{t = h + 1}^{n + h}(Y_t - \hat Y_t)^2}{\frac{1}{n - 1}\sum_{t = 2}^{n }(Y_t -  Y_{t - 1})^2}}.
\end{equation}
After estimating the RMSSE for all the 42,840 time series of the competition, the Weighted
RMSSE (WRMSSE) is used by the organizer for the overall accuracy comparison defined in
\autoref{eq:wrmsse}
\begin{equation}
  \label{eq:wrmsse}
  WRMSSE = \sum_{i = 1}^{42,840} \omega_i \times RMSSE_i
\end{equation}
where $\omega_i$ is the weight of the series based on actual dollar sales of the
product. It is worth mentioning that the actual future values
($Y_{n + 1}, ..., Y_{n + h}$) were not available during the validation frame in the M5
competition.

Interestingly, one may notice that the RMSSE from hierarchical levels of view are equally
weighted, which indicates that the overall accuracy in WRMSSE is primarily affected by the
upper levels forecasts. Because all hierarchical levels are equally weighted with the M5
accuracy metric WRMSSE, it is easier to forecast the continuous time series at upper
levels than to forecast massive intermittent time series at the bottom level.  Based on
the above concern, the hierarchical alignment scheme builds the forecasting strategy
focusing on the forecasting accuracy at the top level forecasts. It aligns them with the
bottom level forecasts.

\autoref{tab:features} shows the time series features used for the LightGBM model at the
bottom level. The feature matrix includes all available categorical features like
\textsf{store\_id} and \textsf{category\_id}. Furthermore, the time since the first sale
of an item, the price, and derived price features (\textsf{price\_min},
\textsf{price\_max}, etc.) are also included.  Events and SNAP data, and time features
like \textsf{day}, \textsf{week} and \textsf{month} are also used in the model.

\begin{table}
  \centering
  \caption{Features for intermittent time series data at the bottom level used in the
    LightGBM.}
  \label{tab:features}
  \resizebox{\textwidth}{!}{
    \begin{tabular}{ll}
      \toprule
      Feature                  & Description                                                                                  \\
      \midrule
      \textsf{item\_id}        & 3049 unique identifiers of items.                                                            \\
      \textsf{dept\_id}        & 16942 unique identifiers of department.                                                      \\
      \textsf{cat\_id}         & 5652 unique identifiers of category, e.g., foods, household, hobbies.                        \\
      \textsf{sell\_price}     & Price of item in store for given date.                                                       \\
      \textsf{event\_type}     & 108 categorical events, e.g. sporting, cultural, religious.                                  \\
      \textsf{event\_name}     & 157 event names for \textsf{event\_type}, e.g. super bowl, valentine's day, president's day. \\
      \textsf{event\_name\_2}  & Name of event feature as given in competition data.                                          \\
      \textsf{event\_type\_2 } & Type of event feature as given in competition data.                                          \\
      \textsf{snap\_CA}        & Binary indicator for SNAP information in CA.                                                 \\
      \textsf{snap\_TX}        & Binary indicator  for SNAP information in TX.                                                \\
      \textsf{snap\_WI}        & Binary indicator for SNAP information in WI.                                                 \\
      \textsf{release}         & Release week of item in store.                                                               \\
      \textsf{price\_max}      & Maximum price for  item in store in the train data.                                          \\
      \textsf{price\_min}      & Minimum price for item in store in the train data.                                           \\
      \textsf{price\_std}      & Standard deviation of price for item in store in the train data.                             \\
      \textsf{price\_mean}     & Mean of price for item in store in train data.                                               \\
      \textsf{price\_norm}     & Normalized sell price. divided by the \textsf{price\_max}.                                   \\
      \textsf{price\_nunique}  & Number of unique prices for item in store.                                                   \\
      \textsf{item\_nunique}   & Number of unique items for a given price in store.                                           \\
      \textsf{price\_diff\_w}  & Weekly price changes for items in store.                                                     \\
      \textsf{price\_diff\_m}  & Price changes of item in store compared to its monthly mean.                                 \\
      \textsf{price\_diff\_y}  & Price changes of item in store compared to its yearly mean.                                  \\
      \textsf{tm\_d}           & Day of month.                                                                                \\
      \textsf{tm\_w}           & Week in year.                                                                                \\
      \textsf{tm\_m}           & Month in year.                                                                               \\
      \textsf{tm\_y}           & Year index in the train data.                                                                \\
      \textsf{tm\_wm}          & Week in month.                                                                               \\
      \textsf{tm\_dw}          & Day of week.                                                                                 \\
      \textsf{tm\_w\_end}      & Weekend indicator.                                                                           \\
      \bottomrule
    \end{tabular}
  }
\end{table}

\autoref{tab:nbeats-par} and \autoref{tab:lgm-par} document the parameter settings for the
N-BEATS model and LightGBM model, respectively. The baseline features for the bottom level
LightGBM model and hyperparameters were taken from a public Kaggle user notebook provided
by Konstantin Yakovlev (\url{https://www.kaggle.com/kyakovlev/m5-simple-fe}). The feature
matrix does not include any time rolling or lagged features and we find the non-time
series features are of great importance for the intermittent time series forecasting with
the LightGBM model.  There are two hyperparameters , (i) the number of epochs for the
N-BEATS model on the top level, and (ii) the optimal loss multiplier for the top-down
alignment. The approach requires a minimal effort of parameter tuning. We use a simple
grid search on the validation window to find the optimal hyperparameters.

\begin{table}
  \centering
  \caption{Parameter settings for N-BEATS ensembles using \texttt{GluonTS}
    \citep{gluonts2020}}
  \label{tab:nbeats-par}
  \resizebox{\textwidth}{!}{
    \begin{tabular}{lp{8cm}r}
      \toprule
      Parameter                         & Description                                                                                                                            & Value               \\
      \midrule
      \texttt{num\_stacks}              & The number of stacks the network should contain.                                                                                       & $30$                \\
      \texttt{widths}                   & Widths of the fully connected layers with ReLu activation in the blocks.                                                               & $512$               \\
      \texttt{meta\_prediction\_length} & Forecast horizon $h$.                                                                                                                  & $28$                \\
      \texttt{meta\_bagging\_size}      & The number of models that share the parameter combination. Each of these models gets a random initialization. & $3$                 \\
      \texttt{meta\_context\_length}    & The number of time units that condition the predictions.                                                                               & $h\times \{3, ~ 5, ~ 7\}$ \\
      \texttt{meta\_loss\_function}     & The loss function (metric) to use for training the models.                                                                             & \texttt{sMAPE}      \\
      \texttt{learning\_rate}           & Learning rate for each boosting round.                                                                                                 & $0.0006$            \\
      \texttt{epochs}                   & The number of epochs used for the optimization algorithm.                                                                              & $\{10, ~ 12\}$                \\
      \texttt{num\_batches\_per\_epoch} & The number of batches in each epoch for the optimization algorithm.                                                                    & $1000$              \\
      \texttt{batch\_size}              & The batch size used in the optimization.                                                                                               & $16$                \\
      \bottomrule
    \end{tabular}
  }
\end{table}

\begin{table}
  \centering
  \caption{Parameter settings for the LightGBM model \citep{ke2017lightgbm}.}
  \label{tab:lgm-par}
  \resizebox{\textwidth}{!}{
  \begin{tabular}{llr}
    \toprule
    Parameter                  & Description                                                               & Value           \\
    \midrule
    \texttt{objective}         & The objective function to maximize with an optimization algorithm         & \texttt{custom} \\
    \texttt{metric}            & Metric(s) to be evaluated on the evaluation set                           & \texttt{rmse}   \\
    \texttt{learning\_rate}    & Learning rate for each boosting round.                                    & $0.2$           \\
    \texttt{lambda\_l1}        & $L_1$ norm penalty to prevent overfitting                                 & $0.5$           \\
    \texttt{lambda\_l2}        & $L_2$ norm penalty to prevent overfitting                                 & $0.5$           \\
    \texttt{bagging\_freq}     & Frequency for bagging at every $k$ iterations                             & $1$             \\
    \texttt{bagging\_fraction} & The proportion of randomly selected  data for the next $k$ iterations.    & $0.85$          \\
    \texttt{colsample\_bytree} & The proportion for randomly selecting a subset of features on each tree.  & $0.85$          \\
    \texttt{colsample\_bynode} & Proportion for randomly selecting a subset of features on each tree node. & $0.85$          \\
    \bottomrule
  \end{tabular}
   }
\end{table}

Finally, in the evaluation phase of the M5 competition, the optimal multiplier value used
on the evaluation time frame was $0.95$ based on a manual grid search in the space of
$(0, ~ 2]$. To produce the final output, we further use the mean ensemble to obtain the
results based on the five closest multipliers [$0.90$, $0.93$, $0.95$, $0.97$, $0.99$] in
the grid around the optimal value of $0.95$.  This ensemble method reduces the high
variance effect on the bottom level forecasts when we apply the LightGBM with the features
used in \autoref{tab:features}. Although the decision-makers focus on the accuracy of top
levels, the low variance forecasts on the bottom level also improve the overall
performance. The extra ensemble step of the five models around the best $\lambda$
mitigates this high variance while preserving the low bias.

\begin{figure}
  \centering
  \includegraphics[width=0.33\textwidth]{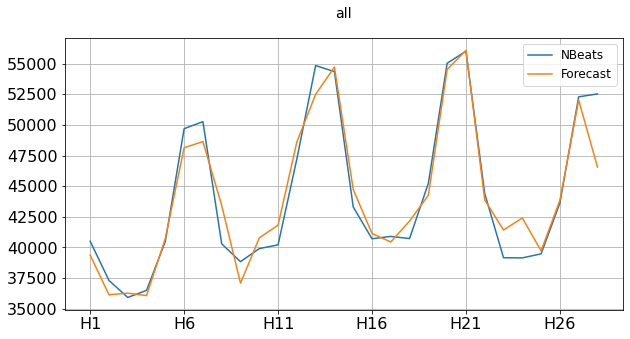}\includegraphics[width=0.33\textwidth]{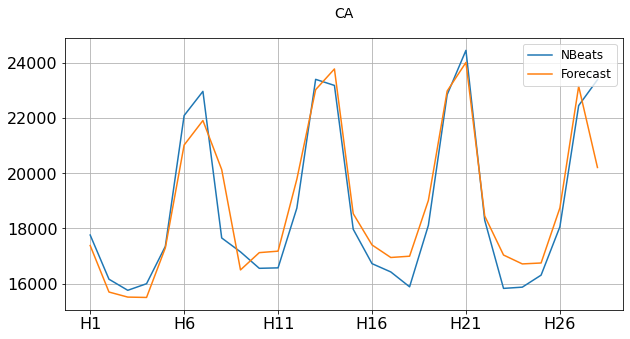}\includegraphics[width=0.33\textwidth]{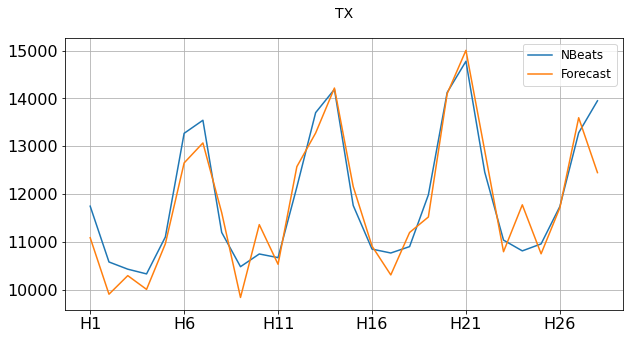} \\
  \includegraphics[width=0.33\textwidth]{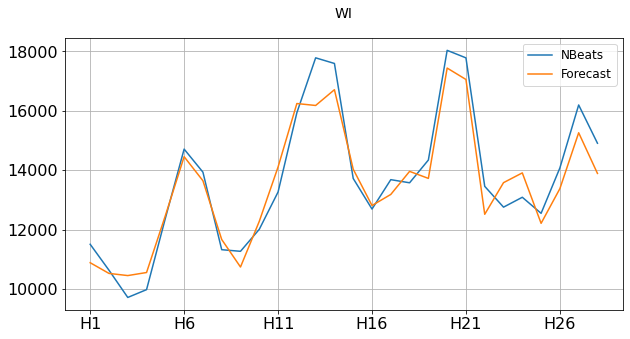}\includegraphics[width=0.33\textwidth]{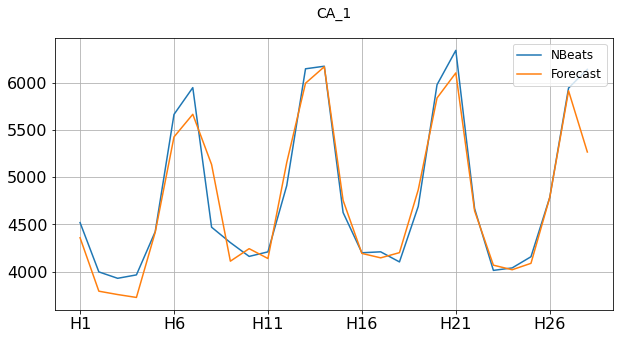}\includegraphics[width=0.33\textwidth]{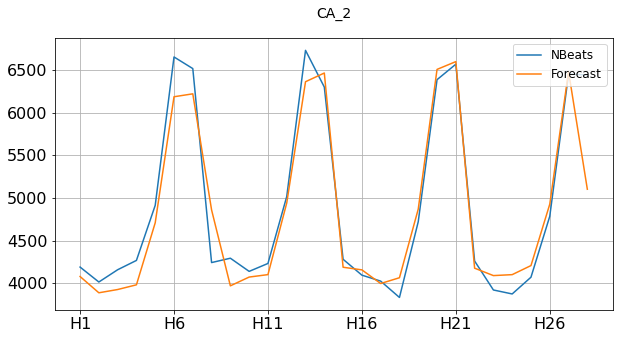} \\
  \includegraphics[width=0.33\textwidth]{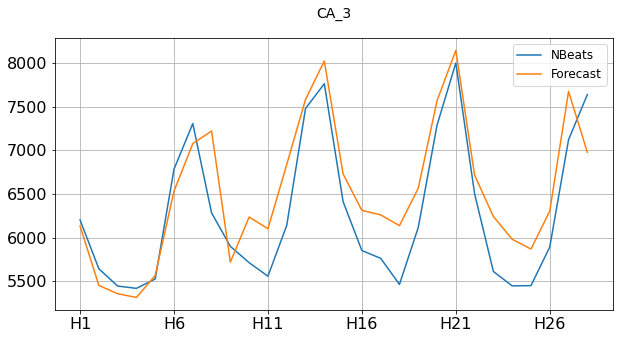}\includegraphics[width=0.33\textwidth]{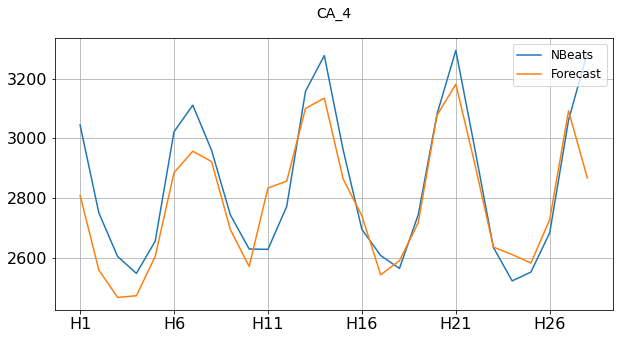}\includegraphics[width=0.33\textwidth]{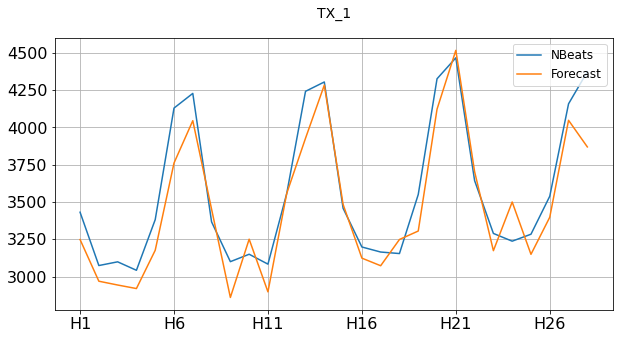} \\
  \includegraphics[width=0.33\textwidth]{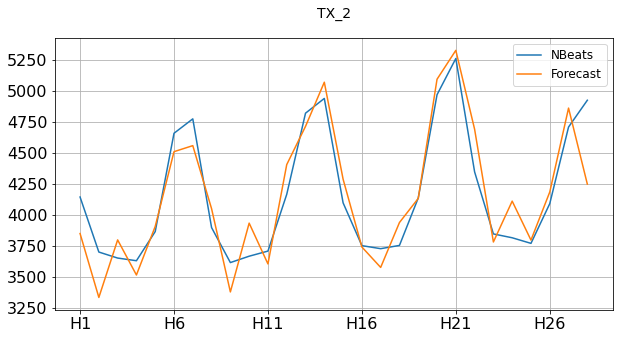}\includegraphics[width=0.33\textwidth]{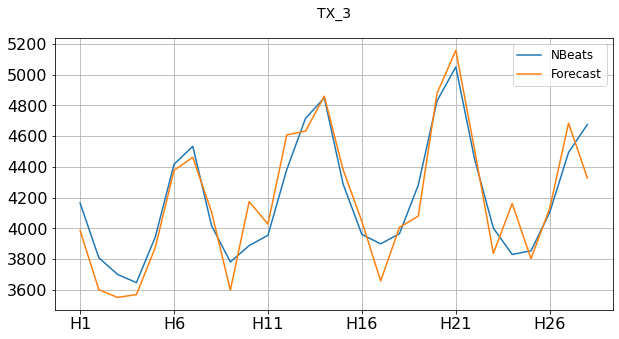}\includegraphics[width=0.33\textwidth]{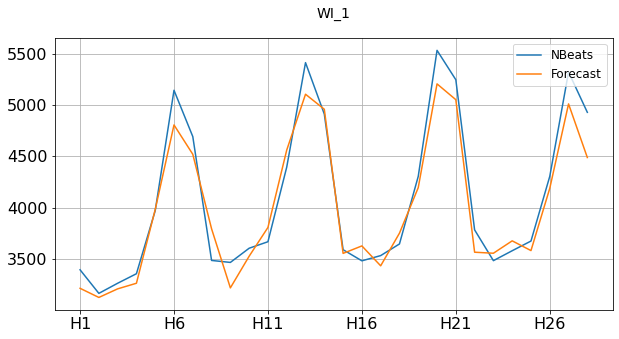} \\
  \includegraphics[width=0.33\textwidth]{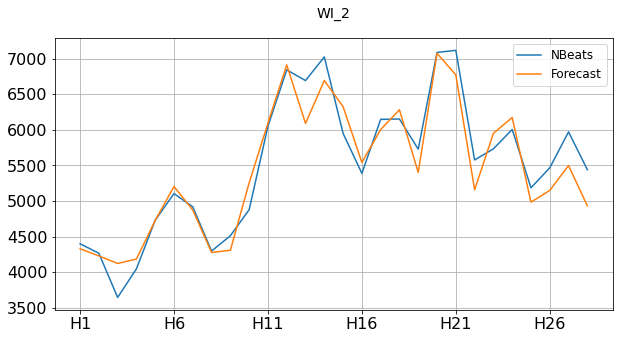}\includegraphics[width=0.33\textwidth]{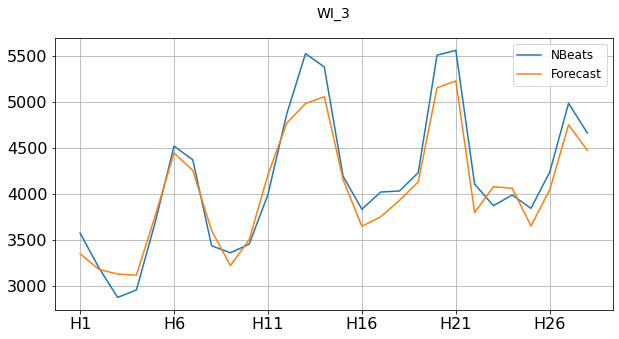}\includegraphics[width=0.33\textwidth]{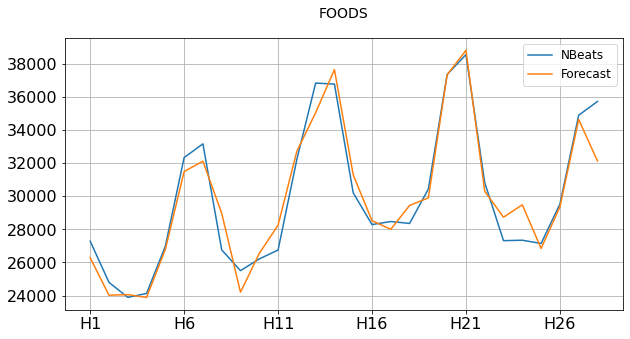} \\
  \includegraphics[width=0.33\textwidth]{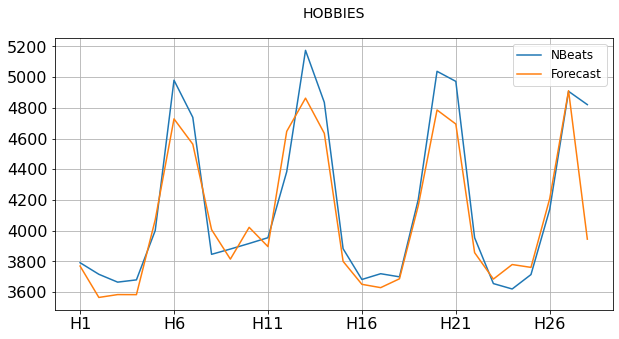}\includegraphics[width=0.33\textwidth]{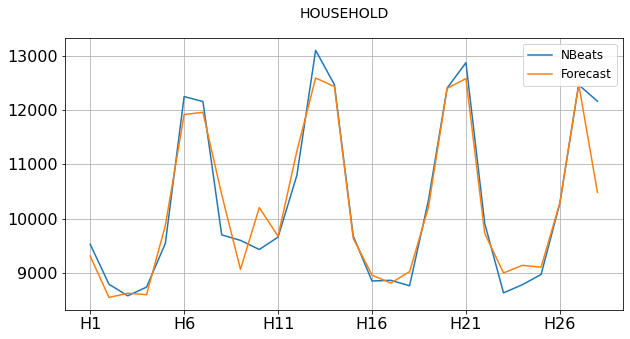}\includegraphics[width=0.33\textwidth]{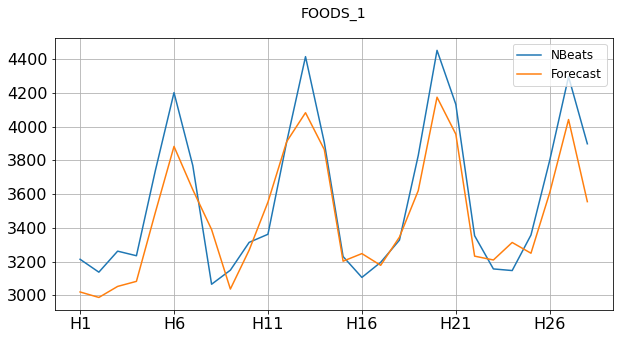} \\
  \includegraphics[width=0.33\textwidth]{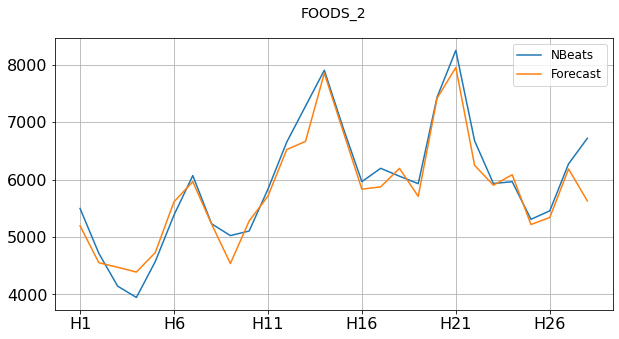}\includegraphics[width=0.33\textwidth]{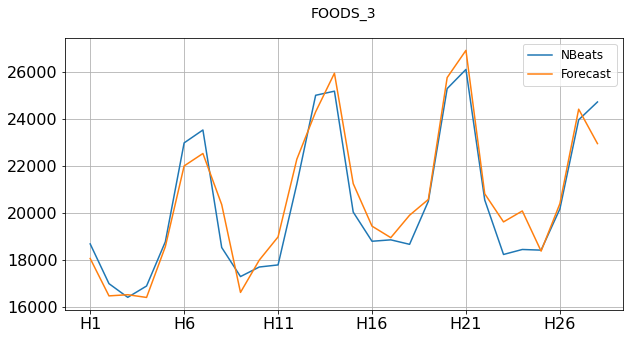}\includegraphics[width=0.33\textwidth]{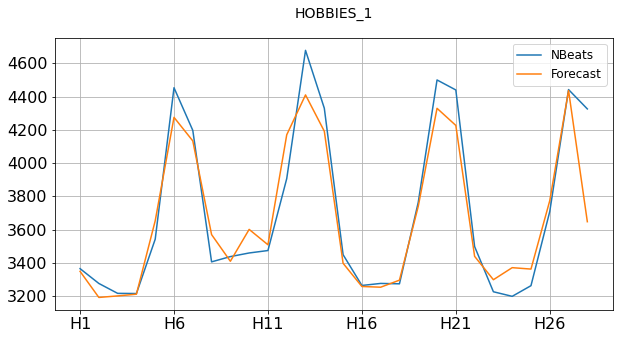} \\
  \includegraphics[width=0.33\textwidth]{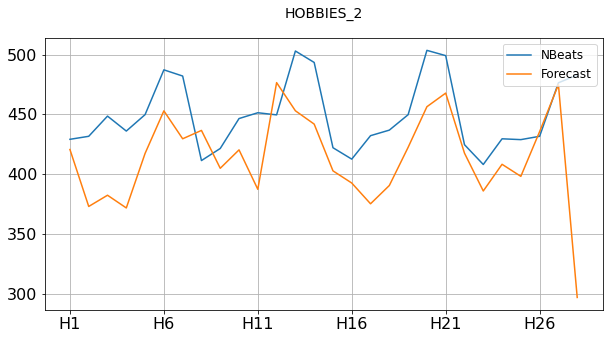}\includegraphics[width=0.33\textwidth]{figures/nbeats_21.png}\includegraphics[width=0.33\textwidth]{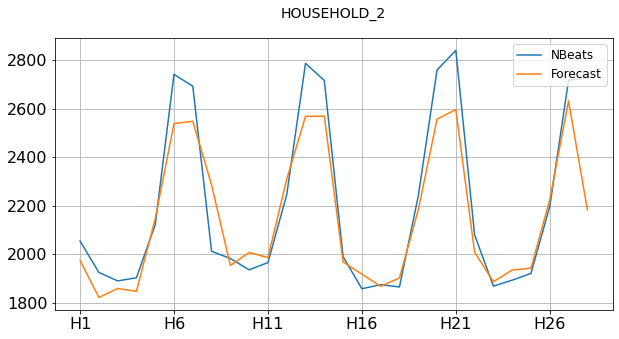} \\
  \caption{Visualizing the N-BEATS forecasts (blue) and the bottom-up aggregated forecasts
    (orange) on top five levels during the evaluation phase of the competition.}
  \label{fig:nbeats}
\end{figure}

During the evaluation frame of the M5 competition, the N-BEATS forecasts and the finally
aggregated forecasts on the top five levels were visualized in
\autoref{fig:nbeats}. \autoref{tab:forecast-error} further depicts the forecasting error
for the aggregated forecasts produced by LightGBM. The top level N-BEATS models are
evaluated with epochs of 10 and 12, respectively.  The mean error was explicitly reported
to check whether the LightGBM has over or under-forecasted the actual values with the
customized gradient in \autoref{eq:gradient}. Specifically, if the forecast is greater
than the actual values, the error will be positive, vise versa.  We ran the model with the
number of epochs from $1$ to $12$. We have noticed that it yields a better overall
forecasting performance with $10$ epochs, which was also used for the final submission of
the M5 competition.

\begin{table}
  \caption{Hierarchical view of forecasting errors for the aggregated forecasts with
    LightGBM. The comparison is based on different accuracy metrics during the evaluation
    phase of the competition. The top level N-BEATS models are evaluated with epochs of 10
    and 12, respectively.}
  \label{tab:forecast-error}
  \sisetup{round-mode=places}
  \centering
  \resizebox{\textwidth}{!}{
    \begin{tabular}{ll S[round-precision=4] S[round-precision=4] l S[round-precision=1]
        S[round-precision=1] l S[round-precision=1] S[round-precision=1] l
        S[round-precision=1] S[round-precision=1] }
  \toprule
           &                       & \multicolumn{2}{c}{RMSSE} &             & \multicolumn{2}{c}{MEAN ERROR} &          & \multicolumn{2}{c}{MAE} &  & \multicolumn{2}{c}{RMSE}                         \\
      \cline{3-4} \cline{6-7}\cline{9-10}\cline{12-13}
 Hierarchy & Category              & \texttt{epochs=10}                        & \texttt{epochs=12}          &                                & \texttt{epochs=10}       & \texttt{epochs=12}                      &  & \texttt{epochs=10}      & \texttt{epochs=12}      &  & \texttt{epochs=10}          & \texttt{epochs=12}          \\
 \midrule
 1         & \textsf{All}          & 8.27424e-05               & 0.00305457  &                                & -50.2829 & 305.514                 &  & 1286.01 & 1379.94 &  & 1786.853358 & 1891.143323 \\
 \cline{2-13}
 2         & \textsf{CA}           & 0.00381747                & 0.00860769  &                                & 172.507  & 259.038                 &  & 774.438 & 776.36  &  & 1011.312975 & 1030.913728 \\
           & \textsf{TX}           & 0.00732117                & 7.66511e-05 &                                & -106.077 & -10.854                 &  & 405.194 & 387.384 &  & 510.311591  & 510.797785  \\
           & \textsf{WI}           & 0.00941973                & 0.0114014   &                                & -179.325 & -197.288                &  & 580.868 & 544.022 &  & 669.370081  & 665.319760  \\
 \cline{2-13}
 3         & \textsf{CA\_1}        & 0.00651318                & 0.0109519   &                                & -62.3624 & -80.8668                &  & 165.797 & 176.687 &  & 251.960114  & 276.002335  \\
           & \textsf{CA\_2}        & 0.00551455                & 2.82492e-06 &                                & -76.8658 & -1.73973                &  & 225.135 & 217.788 &  & 338.398164  & 312.984258  \\
           & \textsf{CA\_3}        & 0.107922                  & 0.156913    &                                & 247.093  & 297.944                 &  & 373.807 & 425.077 &  & 437.354041  & 493.631811  \\
           & \textsf{CA\_4}        & 0.0614284                 & 0.0343661   &                                & -55.3935 & -41.4324                &  & 95.1751 & 99.2969 &  & 128.773389  & 135.573592  \\
           & \textsf{TX\_1}        & 0.0606176                 & 0.00765799  &                                & -108.084 & -38.4165                &  & 156.653 & 126.498 &  & 191.113022  & 162.735428  \\
           & \textsf{TX\_2}        & 8.46274e-06               & 0.00183153  &                                & -1.3454  & 19.7925                 &  & 176.744 & 168.532 &  & 222.483737  & 225.869258  \\
           & \textsf{TX\_3}        & 0.00191014                & 0.00151994  &                                & -14.1593 & 12.6306                 &  & 133.315 & 137.436 &  & 161.767708  & 164.545795  \\
           & \textsf{WI\_1}        & 0.0148716                 & 0.00215254  &                                & -84.4941 & -32.1457                &  & 162.437 & 147.984 &  & 195.600745  & 179.705719  \\
           & \textsf{WI\_2}        & 0.00965457                & 0.0113863   &                                & -64.7727 & -70.3422                &  & 225.545 & 242.784 &  & 282.384300  & 301.882362  \\
           & \textsf{WI\_3}        & 0.0272012                 & 0.0203045   &                                & -101.337 & -87.5526                &  & 186.864 & 143.221 &  & 219.115763  & 179.781174  \\
 \cline{2-13}
 4         & \textsf{FOODS}        & 0.000156301               & 0.00223225  &                                & -44.8391 & 169.452                 &  & 887.706 & 945.779 &  & 1190.379169 & 1237.179938 \\
           & \textsf{HOBBIES}      & 0.0265028                 & 0.00400204  &                                & -80.9121 & -31.4419                &  & 145.731 & 128.193 &  & 220.168978  & 200.773141  \\
           & \textsf{HOUSEHOLD}    & 0.000628359               & 0.00099765  &                                & -35.4896 & 44.7185                 &  & 292.067 & 338.539 &  & 446.830775  & 488.980068  \\
 \cline{2-13}
 5         & \textsf{FOODS\_1}     & 0.0583334                 & 0.0462505   &                                & -99.1094 & -88.2499                &  & 162.945 & 159.121 &  & 191.413724  & 177.728674  \\
           & \textsf{FOODS\_2}     & 0.0255887                 & 0.0223185   &                                & -117.115 & -109.375                &  & 230.504 & 229.946 &  & 320.630416  & 333.228915  \\
           & \textsf{FOODS\_3}     & 0.00836414                & 0.0233191   &                                & 240.441  & 401.471                 &  & 740.073 & 807.622 &  & 900.652408  & 956.991392  \\
           & \textsf{HOBBIES\_1}   & 0.00455652                & 0.0007901   &                                & -31.4341 & 13.0896                 &  & 113.34  & 109.39  &  & 174.904613  & 160.285165  \\
           & \textsf{HOBBIES\_2}   & 1.20657                   & 1.2309      &                                & -35.1637 & -35.5165                &  & 39.1667 & 38.4015 &  & 51.749759   & 52.130153   \\
           & \textsf{HOUSEHOLD\_1} & 0.000279531               & 0.00637765  &                                & 18.2015  & 86.9406                 &  & 224.292 & 264.309 &  & 333.769478  & 366.299706  \\
           & \textsf{HOUSEHOLD\_2} & 0.0267017                 & 0.0174689   &                                & -56.3958 & -45.6152                &  & 103.523 & 96.956  &  & 153.080726  & 150.295519  \\
 \bottomrule
    \end{tabular}
  }
\end{table}

The code is written in Python and reproducible Jupyter Notebooks running on the Kaggle
platform are available at
\url{https://github.com/matthiasanderer/m5-accuracy-competition}.

\section{Ex-post analysis}
\label{sec:postmortem}

The M5 competition used a two-phase testing strategy consisting of the validation and
evaluation phases. In the evaluation phase, no actual future time series values were
available to the public during the competition. After the competition, the optimal value
of the bias multiplier $\lambda$ could be determined by looking at the overall WRMSSE
metric in \autoref{eq:wrmsse}. We notice that the intervals that contain the optimal
multiplier are different in the two phases.

\begin{figure}
  \centering
  \includegraphics[width=0.5\textwidth]{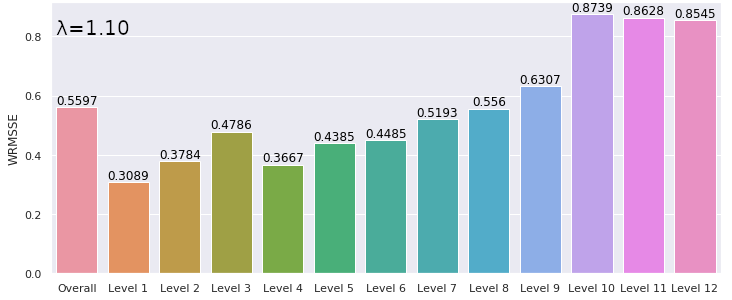}\includegraphics[width=0.5\textwidth]{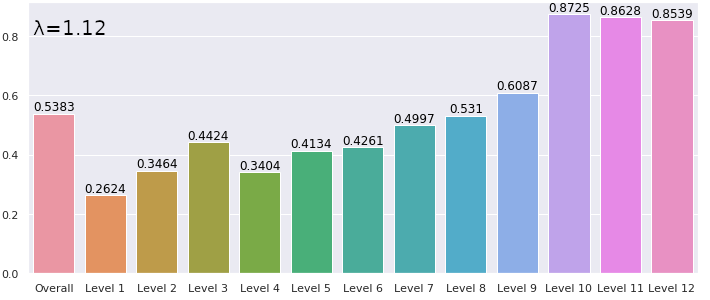} \\
  \includegraphics[width=0.5\textwidth]{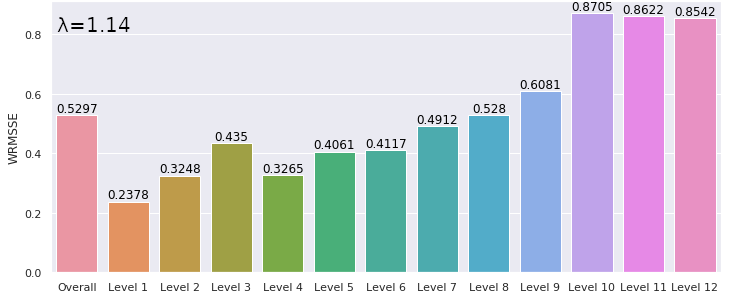}\includegraphics[width=0.5\textwidth]{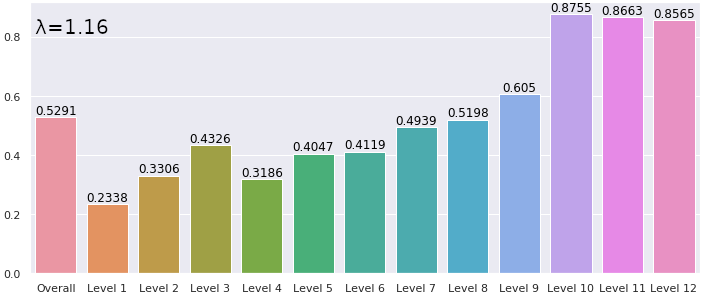} \\
  \includegraphics[width=0.5\textwidth]{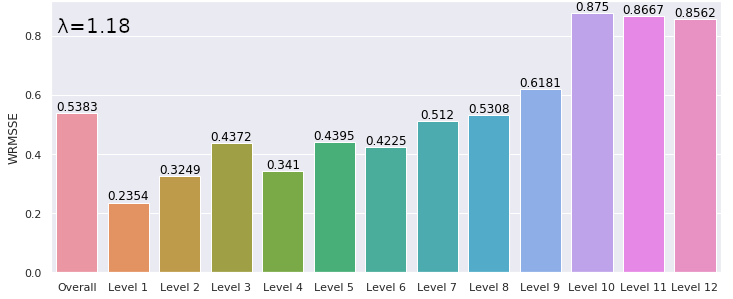}\includegraphics[width=0.5\textwidth]{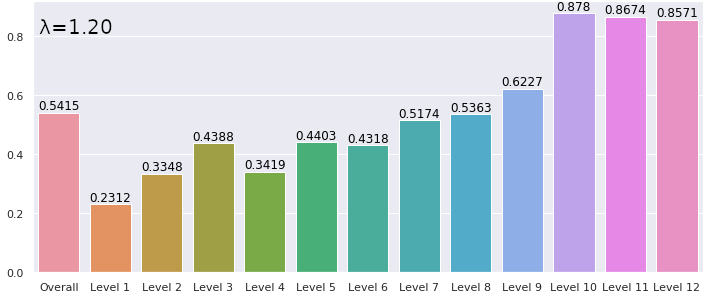}
  \caption{An ex-post analysis for the alignment between forecasting errors and bias
    multiplier in the bottom level model with the validation data of the competition. The
    number of epochs is 10 for the N-BEATS model on the top level.}
  \label{fig:alignment}
\end{figure}

We now check the alignment performance in terms of forecasting errors with different bias
multipliers for the validation time frame with the disclosed actual future
values. \autoref{fig:alignment} shows the forecasting errors (WRMSSE) at all hierarchical
levels based on the bottom level of LightGBM forecasts. One can observe that (i) the lower
levels, which also contain larger portions of data, produce higher forecasting errors
compared to them at upper levels; (ii) the overall WRMSSE on the validation time frame
reaches a minimum of $0.5291$ when $\lambda = 1.16$. The loss multiplier $\lambda$ was
searched in the space of $(0, ~ 2]$, and the WRMSSE is close to the result at the
evaluation time frame with $\mathrm{WRMSSE} = 0.52816$.  This indicates that by
alternating the tuning parameter $\lambda$ in the customized gradient function, one may
lose some accuracy at the bottom level, but the upper levels' accuracy has significantly
improved. As a result, the overall accuracy is improved.

\section{Conclusion}
\label{sec:conclusion}

This paper proposes a \emph{hierarchical-forecasting-with-alignment} approach that focuses
on improving the point forecast accuracy for upper levels by aligning the high accuracy
top level forecasts with the aggregated bottom level forecasts in a hierarchical time
series setting. The proposed top-down alignment approach ensures low forecasting errors on
the upper levels of the hierarchy and improves the overall forecasting performance for an
equally weighted metric like WRMSSE. Our research sheds light on an orthogonal direction
for forecasting reconciliation, as suggested in e.g., \citet{wickramasuriya2019optimal},
allowing some suboptimal forecasts at the lower level while retaining the accuracy on
upper levels.

The hierarchical forecasting with alignment approach is straightforward to implement in
practice. Improving overall forecasting performance requires accurate forecasting on the
top level of the hierarchy. We employ the state-of-the-art deep learning forecasting
approach N-BEATS for continuous time series at the top levels and a widely used tree-based
algorithm LightGBM with non-time series features for the bottom level intermittent time
series. Both methods are easy to use and effortless to scale up with massive time
series. It is worth mentioning that the presented framework is general, and one could
easily replace N-BEATS and LightGBM with other appropriate forecasting algorithms.

One notable difference compared to other approaches in the M5 competition is that the
approach focuses on improving the forecasting accuracy on the continuous upper levels. We
do not directly take the forecasting accuracy from bottom level intermittent time-series
as our central attention, and the bottom level forecasts are treated as mutable to ensure
the hierarchical alignment. However, special combination techniques e.g.,
\citet{kang2021forecast} could improve the accuracy of intermittent time series
forecasting.

Although we focus on the point forecast in this paper, the probabilistic forecast with the
presented scheme should be straightforward to implement with a corresponding probabilistic
loss function.  Another direction for future research is finding the best combination of
top-level and bottom-level models. At the moment, the forecasting for the upper levels and
the bottom level is done independently. To utilize the information across hierarchical
levels, we could consider a joint modeling scheme together with the alignment approach, or
alignment with multiple levels in the future study. Combining the top-down alignment with
other reconciliation methods is also possible but needs further investigation.

\section*{Acknowledgments}

The authors are grateful to the editors, two anonymous reviewers for helpful comments that
improved the contents of the paper. The authors also appreciate the public discussions on
the time series features, models and performance on the Kaggle open forum during the M5
competition. Feng Li is supported by the Beijing Universities Advanced Disciplines
Initiative (No. GJJ2019163), the Emerging Interdisciplinary Project of CUFE and the
disciplinary funding of CUFE.

\printbibliography
\end{document}